\pdfoutput=1

\documentclass[11pt]{article}

\usepackage[]{acl}
\usepackage{times}
\usepackage{latexsym}
\usepackage{subfig}
\usepackage{multirow}
\usepackage{url}
\usepackage{amsfonts}
\usepackage{amssymb}
\usepackage{mathtools}
\usepackage{amsthm}
\usepackage[T1]{fontenc}

\usepackage[utf8]{inputenc}

\usepackage{microtype}

\title{How would Stance Detection Techniques Evolve after the Launch of ChatGPT?}

\author{Bowen Zhang\textsuperscript{1}, Daijun Ding\textsuperscript{1},  Liwen Jing\textsuperscript{2}  \thanks{\ \ Corresponding authors:  ljing@x-institute.edu.cn}\ , Genan Dai\textsuperscript{1}, Nan Yin\textsuperscript{3}\\
  \textsuperscript{1}College of Big Data and Internet, Shenzhen Technology University, Shenzhen, China\\
  \textsuperscript{2}Faculty of Information and Intelligence, Shenzhen X-Institute, Shenzhen, China\\
  \textsuperscript{3}Mohamed bin Zayed University of Artificial Intelligence\\
  }
  
\begin{document}
\maketitle
\begin{abstract}
Stance detection refers to the task of extracting the standpoint (Favor, Against or Neither) towards a target in given texts. Such research gains increasing attention with the proliferation of social media contents. The conventional framework of handling stance detection is converting it into text classification tasks. Deep learning models have already replaced rule-based models and traditional machine learning models in solving such problems. Current deep neural networks are facing two main challenges which are insufficient labeled data and information in social media posts and the unexplainable nature of deep learning models. A new pre-trained language model chatGPT was launched on Nov 30, 2022. For the stance detection tasks, our experiments show that ChatGPT can achieve SOTA or similar performance for commonly used datasets including SemEval-2016 and P-Stance. At the same time, ChatGPT can provide explanation for its own prediction, which is beyond the capability of any existing model. The explanations for the cases it cannot provide classification results are especially useful. ChatGPT has the potential to be the best AI model for stance detection tasks in NLP, or at least change the research paradigm of this field. ChatGPT also opens up the possibility of building explanatory AI for stance detection. 

\footnote{{\color{red} The performance of ChatGPT model in stance detection tasks varies in different released versions. We updated the experimental results to the GPT-3.5-0301 in this updated version of our paper.}}
\end{abstract}

\section{Introduction}
Personal stance towards an issue affects the decision making of an individual, while the stance holds by the public towards thousands of potential topics explains more. Stance detection is an important topic in research communities of both natural language processing (NLP) and social computing \cite{kuccuk2020stance, aldayel2021stance}. Similar to all NLP tasks, early works on stance detection focused on rule-based approaches, and later made a transition into traditional machine learning based algorithms. Since 2014, deep learning models quickly become the mainstream techniques for stance detection. Later on, with the great success of Google's bidirectional encoder representations from transformers (BERT) model, a new NLP research paradigm emerges which is utilizing large pre-trained language models (PLM) together with a fine tuning process. This pre-train and fine-tune paradigm provides exceptional performance for most NLP downstream tasks including stance detection, because the abundance of training data enables PLMs to learn enough general purpose features and knowledge for modeling different languages. Following BERT, more and more PLMs are proposed with different specialties and characteristics, including the ELMo series, the GPT series, the Turing series, varieties of BERT and many more.

ChatGPT is the most recent PLM optimized for dialogue and attracted over 1 million users within 5 days. Programmers use it to interpret code, artists use it to generate prompts for AIGC models, clerks use is to write and translate documents; writers challenge ChatGPT to write poems and film scripts and etc. To what extent will ChatGPT transform the society and people's way of doing and thinking? For NLP experts, is ChatGPT just another pre-trained language model? 

In this work we conduct experiments on ChatGPT for stance detection tasks by directly asking ChatGPT for the result. This approach can be considered as a zero-shot prompting strategy. Experimental results show that ChatGPT can achieve SOTA or similar performance for commonly used datasets including SemEval-2016 and P-Stance with a simple prompt. Since ChatGPT is trained for dialogues, it is surprisingly easy to know the reason of the model's decision making by directly asking why. 
Furthermore, interacting with ChatGPT with a chain of inputs can potentially further improves the performance. This paper is structured as follows: after a brief overview of related work in Section 2, our proposed prompting methods and results are detailed in Section 3. Section 4 contains discussions and future work. 

\section{Related Work}
Before getting into more detail, we first give a formal definition of stance detection. For an input in the form of a piece of text and a target pair, stance detection is a classification problem where the stance of the author of the text is sought in the form of a category label from this set: \{Favor, Against, Neither\}. 
Occasionally, the category label of Neutral is also added to the set of stance categories and the target may or may not be explicitly mentioned in the text.
Researchers approach this task by converting it into a text classification task.
Stance detection studies originally focused on parliamentary debates and gradually shifted to social media contents including Twitter, Facebook, Instagram, online blogs and etc. The techniques to approach these problems also evolve with time. 

Early research works on stance detection from the 1950s mainly adopted rule-based techniques \cite{anand2011cats, walker2012stance}. 
Since the 1990s, machine learning based models gradually replaced small scale rule-based methods. 
Traditional machine learning models build text classifiers for stance detection based on selected features.
The effective algorithms for the classifiers are support vector machine (SVM) \cite{addawood2017stance, mohammad2017stance}, logistic regression \cite{ferreira2016emergent, tsakalidis2018nowcasting, skeppstedt2017detection}, naive bayes \cite{hacohen2017stance,simaki2017stance}, decision tree \cite{wojatzki2016stance} and etc.
With the fast advancement of deep learning in the 2010s, models based on deep neural networks (DNN) become mainstream in this field.
These methods design neural networks with different structures and connections to obtain the desired stance classifier, which can be categorized as conventional DNN models, attention-based DNN models and graph convolutional network (GCN) models.  
Convolutional neural network (CNN) and long short-term memory (LSTM) models are most commonly used conventional DNN models \cite{augenstein2016stance, du2017stance};
the attention-based methods mainly utilize target-specific information as the attention query, and deploy an attention mechanism for inferring the stance polarity \cite{dey2018topical, sun2018stance};
and the GCN methods propose a graph convolutional network to model the relation between target and text \cite{LiPLSLWYH22, bowenacl, ConfortiBPGTC21}.

Inspired by the recent success of PLMs, fine-tuning methods have led to improvements in stance detection tasks\cite{liu2021enhancing}.
Fine-tuning models adapt PLMs by building a stance classification head on top of the ``{$<$cls$>$}'' token, and fine-tune the whole model. The PLMs are getting larger and larger because the performance and sample efficiency on downstream tasks are normally proportional to the scale of the model, and some abilities like the prompting strategies, popularized by GPT-3, are considered to be effective only when the model reaches a certain scale \cite{wei2022emergent}.
The main idea of prompt-based methods is mimicking PLMs to design a template suitable for classification tasks and then build a mapping (called verbalizer) from the predicted token to the classification labels to perform class prediction, which bridges a projection between the vocabulary and the label space. The prompting strategies provide further improvements for stance detection performance\cite{shin2020autoprompt}. Models like LaMDA, GPT-3 and etc. also gain success on few-shot prompting\cite{wei2022emergent}, which alleviates the demand for large amount of training data and the tedious training process.

Generally speaking, stance detection techniques and NLP algorithms in general experienced four main paradigms: (1) rule-based models; (2) traditional machine learning based models; (3) deep neural network models and (4) PLM pre-train and fine-tune paradigm. Quite recently, the 5th paradigm "pre-train, prompt and predict" starts to draw wide attention\cite{liu2021pre}.

\begin{table}[h!]
\small
\begin{center}
\begin{tabular}{lccc}
\hline
Model    & HC   & FM   & LA   \\ \hline
Bicond$^\dag$ \cite{augenstein2016stance}  & 32.7 & 40.6 & 34.4 \\
CrossNet$^\dag$ \cite{xu2018cross} & 38.3 & 41.7 & 38.5 \\
SEKT  \cite{zhang2020enhancing}   & 50.1 & 44.2 & 44.6 \\
TPDG$^\dag$  \cite{LiangF00DHX21}   & 50.9 & 53.6 & 46.5 \\
Bert\_Spc$^\dag$ \cite{Bert}     & 49.6 & 41.9 & 44.8 \\
Bert-GCN$^\dag$ \cite{linbertgcn} & 50.0 & 44.3 & 44.2 \\
PT-HCL$^\dag$ \cite{liang2022zero}  & 54.5 & 54.6 & 50.9 \\ \hline
ChatGPT  &   \textbf{78.0}    &  \textbf{69.0}  &   \textbf{59.3}  \\ \hline 
\end{tabular}
\end{center}
\caption{Performance comparison (F1-avg) on SemEval-2016 dataset with zero shot setup.}
\label{tab1}
\end{table}

\begin{figure*}[htbp]
	\centering
	\includegraphics[width=0.85\linewidth]{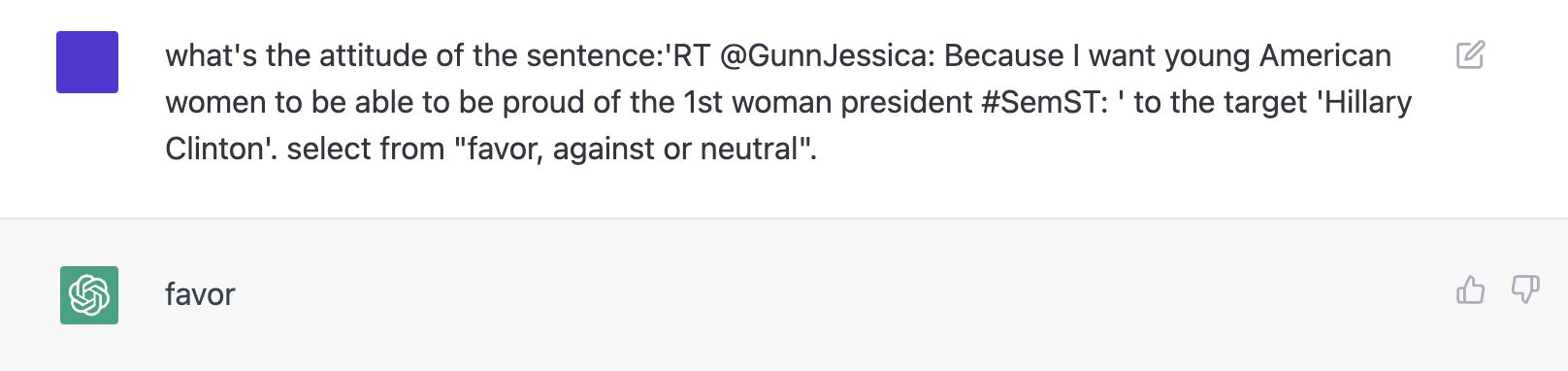}
	\caption{Example of Question to ChatGPT}
 	\label{fig1}
\end{figure*}

\begin{table*}[h!]
\small
\begin{center}
\begin{tabular}{lllllll}
\hline
Methods           & \multicolumn{2}{c}{FM}         &  \multicolumn{2}{c}{LA}        &    \multicolumn{2}{c}{HC}    \\ \cline{2-7}
                  & F1-m & F1-avg & F1-m & F1-avg & F1-m & F1-avg   \\ \hline
BiLSTM \cite{augenstein2016stance}            & 48.0   & 52.2     & 51.6   & 54.0     & 47.5   & 57.4        \\
BiCond \cite{augenstein2016stance}            & 57.4   & 61.4     & 52.3   & 54.5     & 51.9   & 59.8         \\
TextCNN  \cite{kiml}         & 55.7   & 61.4     & 58.8    & 63.2      & 52.4   & 58.5         \\
MemNet   \cite{TangQL16}         & 51.1   & 57.8     & 58.9   & 61.0     & 52.3    & 60.3         \\
AOA   \cite{huang2018aspect}            & 55.4   & 60.0     & 58.3   & 62.4     & 51.6   & 58.2        \\
TAN    \cite{du2017stance}            & 55.8   & 58.3     & 63.7   & 65.7     & 65.4   & 67.7        \\
ASGCN   \cite{zhang2019aspect}          & 56.2   & 58.5     & 59.5   & 62.9     & 62.2   & 64.3         \\
Bert\_Spc  \cite{Bert}       & 57.3    & 60.6      & 64.0   & 66.3    & 65.8   & 69.1         \\
TPDG  \cite{LiangF00DHX21}            & 67.3    & /        &     74.7  &  /       & 73.4   &  /             \\ \hline
ChatGPT  & \textbf{68.4}   & \textbf{69.0}    & 58.2   & 59.3    & \textbf{79.5 }  & \textbf{78.0}     
\\
					
\hline
\end{tabular}
\end{center}
\caption{Performance comparison on SemEval-2016 dataset with in-domain setup.}
    \label{tab2}
\end{table*}

\begin{table*}[h!]
\small
    \centering

\begin{tabular}{lllllll}
\hline
Methods           & \multicolumn{2}{c}{Trump}         &  \multicolumn{2}{c}{Biden}        &    \multicolumn{2}{c}{Bernie}    \\ \cline{2-7}
                  & F1-m & F1-avg & F1-m & F1-avg & F1-m & F1-avg   \\ \hline
BiLSTM  \cite{augenstein2016stance}              & 69.7  & 72.0     & 68.7  & 69.5       & 63.8      & 63.9   \\
BiCond  \cite{augenstein2016stance}              & 70.6  & 73.0      & 68.4  & 69.4     & 64.1   & 64.6    \\
TextCNN   \cite{kiml}           & 76.9   & 77.2       & 78.0    & 78.2     & 69.8     & 70.2     \\
MemNet  \cite{TangQL16}         & 76.8  & 77.7    & 77.2   & 77.6        & 71.4    & 72.8      \\
AOA  \cite{huang2018aspect}        & 77.2    & 77.7   & 77.7   & 77.8    & 71.2      & 71.7     \\
TAN      \cite{du2017stance}           & 77.1  & 77.5    & 77.6   & 77.9    & 71.6       & 72.0      \\
ASGCN   \cite{zhang2019aspect}      & 76.8  & 77.0      & 78.2   & 78.4     & 70.6    & 70.8      \\
Bert\_Spc   \cite{Bert}           & 81.4  & 81.6       & 81.5  & 81.7   & 78.3   & 78.4            \\ \hline
ChatGPT    & \textbf{82.8}  & \textbf{83.2}   & \textbf{82.3}   & \textbf{82.0 }   & \textbf{79.4 }  & \textbf{79.4}     
\\
					
\hline
\end{tabular}
    \caption{Performance comparison on P-Stance dataset with in-domain setup.}
        \label{tab3}
\end{table*}

\section{Methods and Results}

\textbf{Task definition:}
We use $X = \{x, p\}_{i=1}^{N}$ to denote the collection of data, where each $x$ denotes the input text and $p$ denotes the corresponding target. $N$ represents the number of instances.
Stance detection aims to predict a stance label for the input sentence $x$ towards the given target $p$ by using the stance predictor.

In this Section, we reveal the performance of the ChatGPT method for stance detection.
We utilize a special case of prompt to construct the stance predictor by creating a template of a direct question.
Specifically, we directly ask the ChatGPT model the stance polarity of a certain tweet towards a specific target.
Figure \ref{fig1} shows an example. Given the input: ``\textit{RT \@GunnJessica: Because i want young American women to be able to be proud of the 1st woman president \#SemST}'',  the question for ChatGPT input is: \textit{``What is the attitude of the sentence : "RT \@GunnJessica: Because i want young American women to be able to be proud of the 1st woman president \#SemST'' to the target ``Hillary Clinton" select from ``favor, against' or neutral'.} For this particular example, ChatGPT returns a correct result.


\begin{figure*}[h!]
	\centering
	\includegraphics[width=0.85\textwidth]{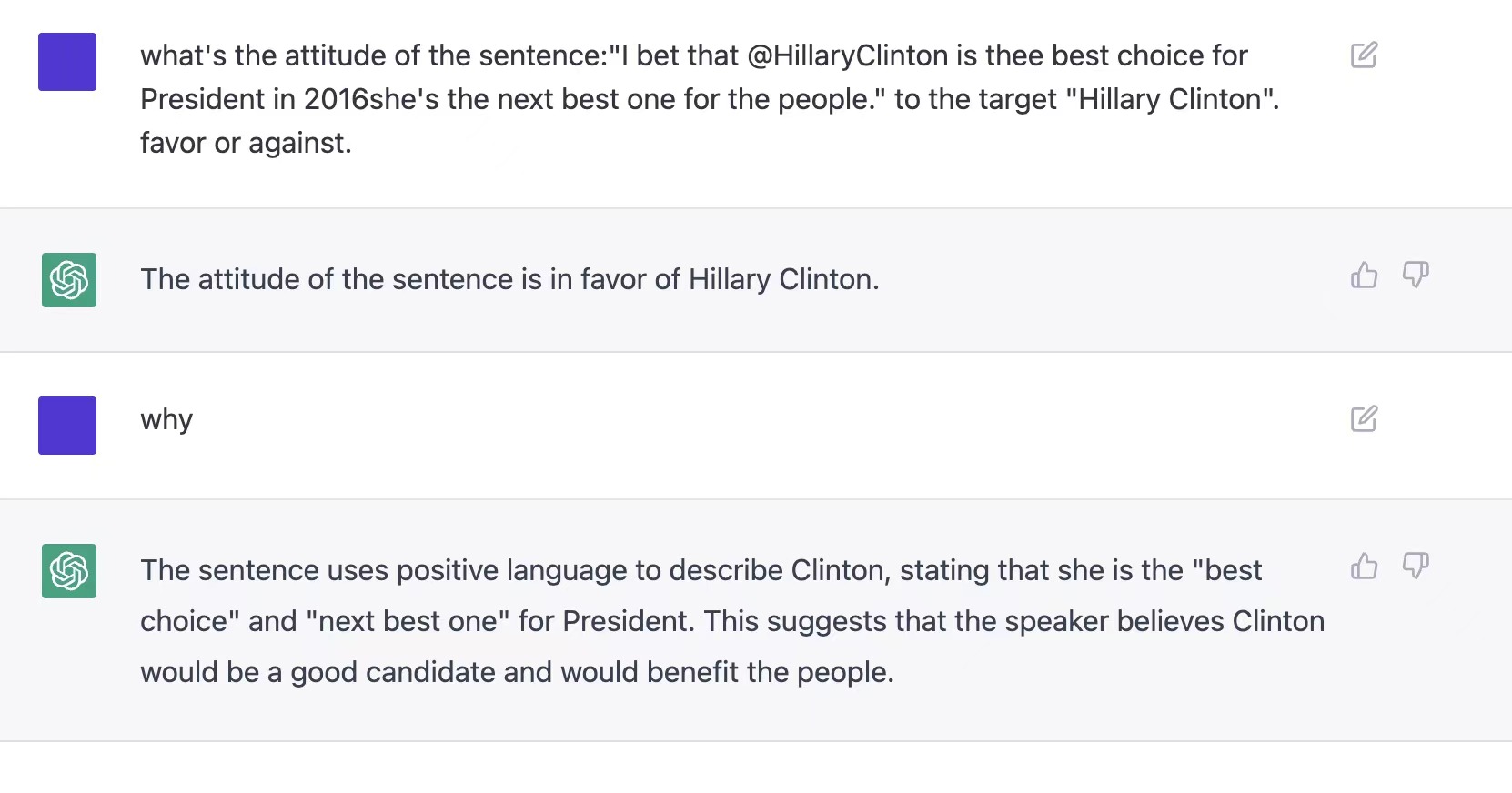}
	\caption{ChatGPT's Explanation when the Stance is Explicitly Expressed in the Text}
	\label{fig2}
\end{figure*}

\begin{figure*}[h!]
	\centering
	\includegraphics[width=0.85\textwidth]{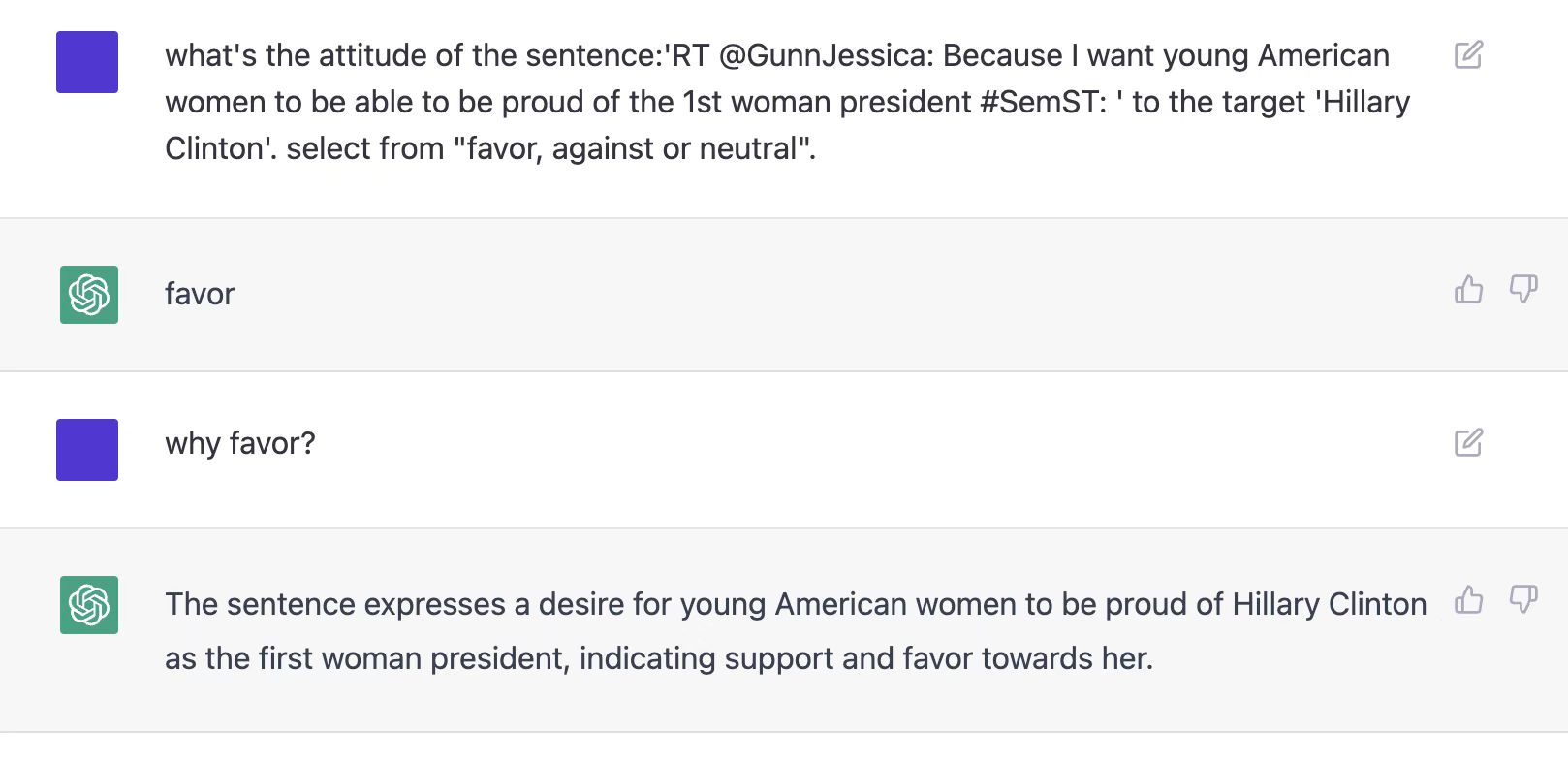}
	\caption{ChatGPT's Explanation when the Stance is Implicitly Expressed in the Text}
	\label{fig3}
\end{figure*}

\textbf{Results:}
To compare the effectiveness of ChatGPT, we carried out experimental validations in the SemEval-2016 stance dataset\cite{StanceSemEval2016} and P-Stance dataset\cite{li2021p}. The SemEval-2016 is a dataset of 4870 tweets in English with manual annotation for stance towards 6 selected targets and ‘Hillary Clinton (HC)’,
‘Feminist Movement (FM)’ and ‘Legalization of Abortion (LA)' are three commonly used ones. 
Similarly, P-Stance dataset contains 21574 English tweets with political contents with stance annotations towards three targets including “Donald Trump,” “Joe Biden,” and
“Bernie Sanders.” (Note that, following \cite{li2021p}, we only use favor and against labels for evaluation.)
Since OpenAI has not provide API for using ChatGPT yet, experiments has only been conducted on these two benchmark datasets on social media texts.
Following \cite{zhang2020enhancing, liang2022zero} we use the F1-avg (the average of F1 on Favor and Against), and macro-F1 score (denoted as F1-m) for performance evaluation.

We constructed both zero-shot stance detection and in-domain stance detection setups for results comparison. The zero-shot setup means the model is directly tested without any adjustment with training data, which is a fair comparison with our proposed prompt method using ChatGPT. We also compared our zero-shot results of ChatGPT with other mainstream stance detection models in an in-domain setup, which means these models are optimized with 80\% tweets as training data. The 
results are summerized in Table \ref{tab1} to \ref{tab3}.

The results show that ChatGPT achieves SOTA results in zero-shot setup.
For example, ChatGPT achieves a 16.6\% improvement on average compared with the best competitor PT-HCL in zero-shot setup.
Compared with in-domain setup, where these methods first learned from 80\% training corpus, ChatGPT still yields better performance than all the baselines in most tasks.

\begin{figure*}[h!]
	\centering
	\includegraphics[width=0.85\textwidth]{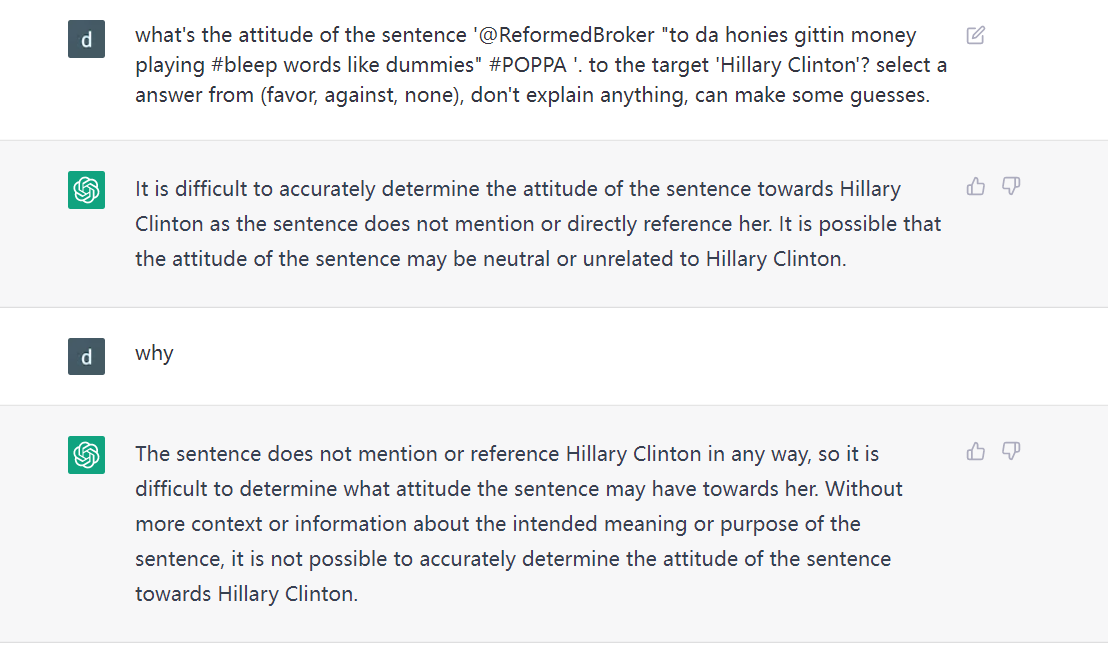}
	\caption{ChatGPT's Explanation when it cannot Provide a Stance Detection Result (Case 1)}
	\label{fig4}
\end{figure*}

\begin{figure*}[h!]
	\centering
	\includegraphics[width=0.85\textwidth]{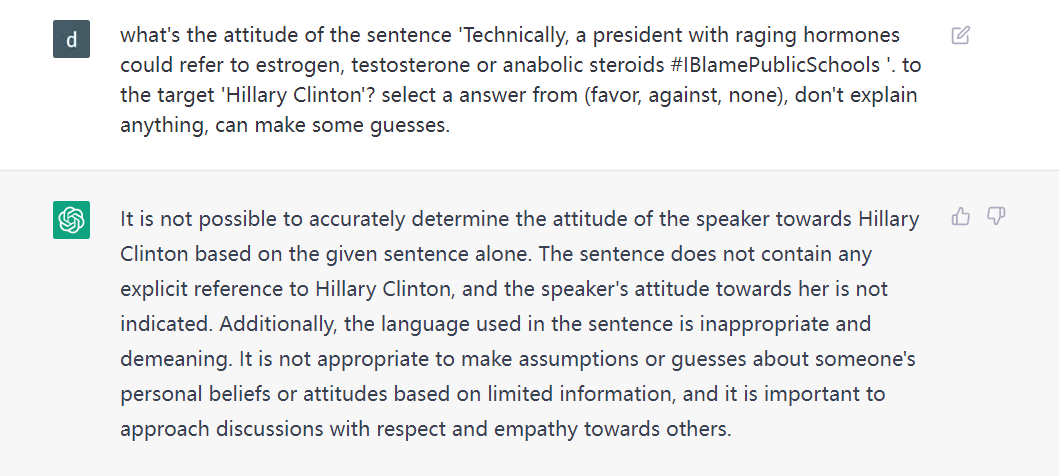}
	\caption{ChatGPT's Explanation when it cannot Provide a Stance Detection Result (Case 2)}
	\label{fig5}
\end{figure*}

\section{Discussions and Future Work}
Results in Section 3 demonstrate the emergent ability of ChatGPT on zero-shot prompting for stance detection tasks. By using a simple prompt of directly asking the dialogue model for the stance with no training, ChatGPT returns SOTA results in both zero-shot and in-domain setups. 
The launch of ChatGPT would potentially transform the whole research area. We would like to discuss three research directions which might further improve the performance of ChatGPT on stance detection tasks.

(1) \textbf{Are there better prompt templates?}

In this work, only one prompt template for stance detection has been tested with ChatGPT. Engineering the prompt template may further improve the zero-shot performance of using ChatGPT or unlock the use of ChatGPT to other NLP tasks. Futher studies can take the intuitive approach of manually selecting prompt templates or design an automated process for template selection.

(2) \textbf{How well can ChatGPT explain itself?}

ChatGPT is a language model trained for dialogues, thus it is a natural next step to ask the model why it provides certain answer. As shown in Figure \ref{fig2} and \ref{fig3}, ChatGPT provides perfect explanations for why the given tweet is in favor of the target Hillary Clinton weather the stance is explicitly or implicitly expressed in the text. 
Such results indicate that ChatGPT carries out stance classification based on logic reasoning instead of pure probability calculation.
These explanations opens up the possibility of building explanatory AI for stance detection.

(3)\textbf{Can multi-round conversation help to improve the results?}

ChatGPT has already shown exceptional results with zero-shot prompting, however, it is more powerful than a stance classifier. For some instances when ChatGPT cannot provide prediction results, it can still explain why it cannot produce a prediction, e.g. "the sentence does not mention or directly reference the target", as shown in \ref{fig4} or even "instruct the speaker to express opinion with respect and empathy", as shown in \ref{fig5}. These explanations help us select the innately flawed data in the dataset, for which no model and even no human can accurately decide the stance only by the given information. For those flawed tweets, it is still possible to determine the stance of it by fixing the issue in the following conversation. In a multi-round conversation with ChatGPT, we can feed a variety of information to the model including background knowledge, missing part of the sentence, stance classification examples and etc. Future investigation on how to design a multi-round conversation may further improve the performance of ChatGPT model on more NLP tasks including stance detection.

\bibliography{anthology,custom}
\bibliographystyle{acl_natbib}




\end{document}